\documentclass[journal]{IEEEtran}

\usepackage{amsmath,amsfonts,amssymb,amsbsy,fixmath,eucal}
\usepackage{cite}
\usepackage{algorithmic}
\usepackage{graphicx}
\usepackage{textcomp}
\usepackage[caption=false]{subfig}
\usepackage{float}
\usepackage{times}
\usepackage[utf8]{inputenc}
\usepackage{graphicx}
\usepackage{amsthm}
\usepackage{booktabs}
\usepackage{url}
\usepackage{xr}
\usepackage{stfloats}
\usepackage{multirow}

\begin{document}

\title{Anomaly Detection in Time Series with Triadic Motif Fields and Application in Atrial Fibrillation ECG Classification}
\author{Yadong Zhang and Xin Chen*
\thanks{This work was supported in part by the National Natural Science Foundation of China (Grant No.21773182 (B030103)) and the HPC Platform, Xi'an Jiaotong University. Xin Chen thanks Prof. Chunhua Shen at the University of Adelaide for the discussion on the transfer learning and feature extraction. (Corresponding author: Xin Chen.)}
\thanks{Yadong Zhang is with Center of Nanomaterials for Renewable Energy, School of Electrical Engineering, Xi'an Jiaotong University, Xi'an, Shaanxi, China (e-mail: zhangyadong@stu.xjtu.edu.cn). }
\thanks{Xin Chen is with Center of Nanomaterials for Renewable Energy, School of Electrical Engineering, Xi'an Jiaotong University, Xi'an, Shaanxi, China (e-mail: xin.chen.nj@xjtu.edu.cn). }
}

\maketitle

\begin{abstract}
In the time-series analysis, the time series motifs and the order patterns in time series can reveal general temporal patterns and dynamic features. Triadic Motif Field (TMF) is a simple and effective time-series image encoding method based on triadic time series motifs. Electrocardiography (ECG) signals are time-series data widely used to diagnose various cardiac anomalies. The TMF images contain the features characterizing the normal and Atrial Fibrillation (AF) ECG signals. Considering the quasi-periodic characteristics of ECG signals, the dynamic features can be extracted from the TMF images with the transfer learning pre-trained convolutional neural network (CNN) models. With the extracted features, the simple classifiers, such as the Multi-Layer Perceptron (MLP), the logistic regression, and the random forest, can be applied for accurate anomaly detection. With the test dataset of the PhysioNet Challenge 2017 database, the TMF classification model with the VGG16 transfer learning model and MLP classifier demonstrates the best performance with the 95.50\% ROC-AUC and 88.43\% F1 score in the AF classification. Besides, the TMF classification model can identify AF patients in the test dataset with high precision. The feature vectors extracted from the TMF images show clear patient-wise clustering with the t-distributed Stochastic Neighbor Embedding technique. Above all, the TMF classification model has very good clinical interpretability. The patterns revealed by symmetrized Gradient-weighted Class Activation Mapping have a clear clinical interpretation at the beat and rhythm levels.
\end{abstract}


\section{Introduction}
\label{sec:introduction}
\IEEEPARstart{T}{ime} series classification and anomaly detection are important techniques in the understanding of the varieties of dynamics in Science and Engineering \cite{TSC_review_2009,anomaly_2009}. The temporal patterns in time series contain important information about the underlying dynamics. The electrocardiogram (ECG) is the important medical time series used by cardiologists and medical practitioners for monitoring cardiac health and detecting cardiac abnormality, such as Atrial Fibrillation (AF), Myocardial Infarction (MI), {\it etc}. 
ECG signals from normal healthy and abnormal hearts have certain clinical patterns. Identifying and learning these temporal patterns/features in the ECG time series are critically essential to the ECG analysis \cite{ECG_review_2017}.  

The conventional ECG anormaly detection and signal classification require the identification of the AF features according to the expert rules in the ECG diagnosis \cite{RR_traditional_2001,AF_detection_2017}. The random forest classifiers \cite{AF_detection_2017} are implemented with the features from time, frequency, time-frequency domain and phase space reconstruction. 
Recently, deep learning is successfully applied in the ECG analysis \cite{1D_CNN_2016,LSTM_ECG_2019,CRNN_ECG_2017,mina_2019}. Feature extraction is carried out automatically with deep learning models. 
Convolutional Neural Networks \cite{1D_CNN_2016}, BiLSTM-Attention Neural Network \cite{LSTM_ECG_2019} and Convolutional Recurrent Neural Networks \cite{CRNN_ECG_2017}, {\it etc.} are used to learn temporal patterns and features of the ECG signals in the time and frequency domain. MultIlevel kNowledge-guided Attention networks \cite{mina_2019} is implemented with multilevel attention in the knowledge of beat, rhythm, and frequency information in the AF classification. 

Motivated by the success of Convolutional Neural Network (CNN) in the image classification, the time series classification with image encoding of time series also demonstrates high performance, {\it{e.g.}} Gramian Angular Summation/Difference Fields (GASF/GADF) and Markov Transition Fields (MTF) \cite{GAF_MTF_2015}, the Recurrence plots \cite{Recurrence_plot} and time-frequency-domain images \cite{CRNN_ECG_2017}. The Tiled CNN can classify the time series based on the GASF/GADF images \cite{GAF_MTF_2015}. CNN is applied for the classification of time series based on the Recurrence plots \cite{Recurrence_plot}. CRNN based on spectrogram images of ECG signals do well in the AF classification \cite{CRNN_ECG_2017}.

Time series motifs define the local order patterns in time series. Motif occurrence probabilities provide the information about the complexity of the underlying dynamics. Permutation entropy \cite{PE_2002} according the motif ordinal patterns has been successfully used for the time series complexity measurement \cite{PE_2002}, chaotic maps characterization \cite{chaotic_PE_2013}, stock market analysis \cite{stock_PE_2009}, ECG signals analysis \cite{ECG_PE_2015}, and {\it etc}. The statistical probabilities of triadic time series motif have been applied for the UCR time series archive classification \cite{triadic_classification_2019}.

In this paper, a novel image encoding method, Triadic Motif Field (TMF) is proposed by considering the triadic motifs in time series. With the TMF images, we successfully detect AF ECG signals with transferred knowledge from ImageNet dataset. And the clinical patterns can be interpreted using symmetrized Gradient-weighted Class Activation Mapping (Grad-CAM). The proposed method is simple, effective, and accurate. Above all, it has good interpretability compared to other AF classification methods based on deep learning.
Section~\ref{sec:method} discusses the TMF image encoding method and describes the feature extraction with the TMF images and the classification model. In Section~\ref{sec:benchmark}, the performance of the classification model is evaluated and discussed. In Section~\ref{sec:interpret}, the interpretability of the TMF  classification model is discussed. 

\section{Method}
\label{sec:method}
\begin{figure}[H]
  \centering
  \includegraphics[width=0.5\textwidth]{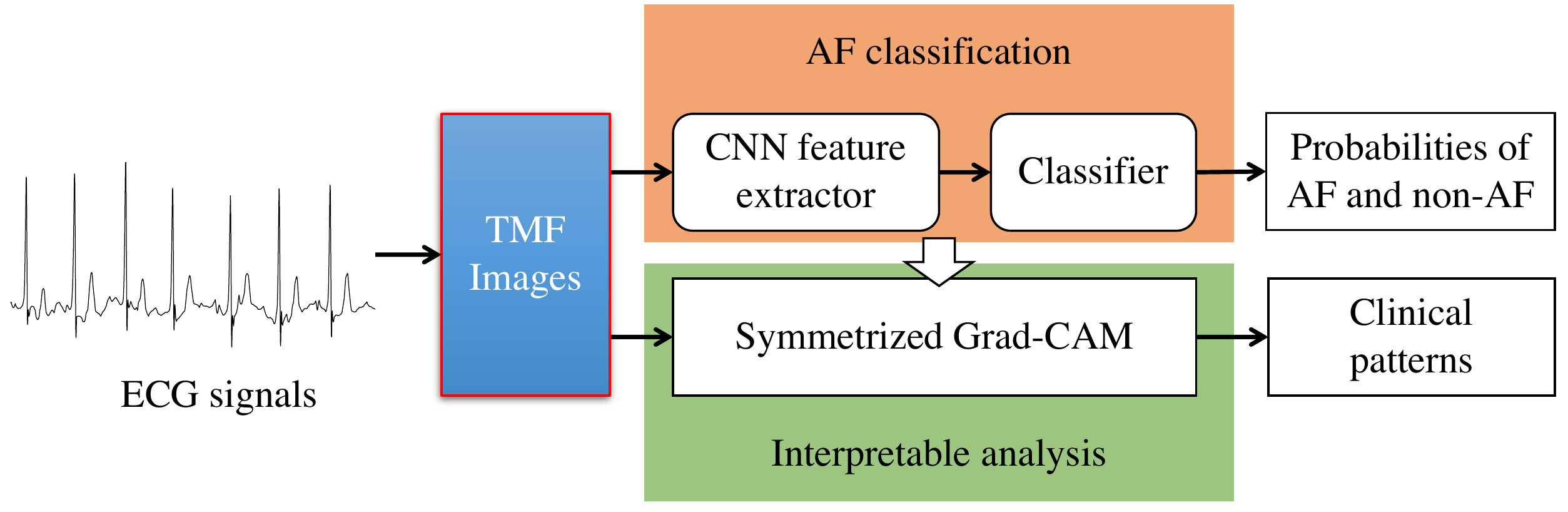}
  \caption{The framework of interpretable classification model based on the TMF images. The probabilities of AF and non-AF are predicted with feature transfer learning. On the parallel, symmetrized Grad-CAM is used for the pattern recognitions in the AF and non-AF ECG signals.}
  \label{fig:method}
  \end{figure}

Fig.~\ref{fig:method} shows the image encoding method, Triadic Motif Field (TMF), that converts the order patterns and temporal structures of time series to images based on triadic motifs in ECG signals. Using the transfer learning pre-trained CNN models based on the ImageNet dataset, the TMF images are used to extract the features for the Atrial Fibrillation (AF) and non-AF ECG signals. The symmetrized Gradient-weighted Class Activation Mapping (Grad-CAM) is used to interpret results by identifying beat-level and rhythm-level patterns. 

\subsection{Triadic Motif Field Images}
\begin{figure}[H]
\centering
\includegraphics[width=0.5\textwidth]{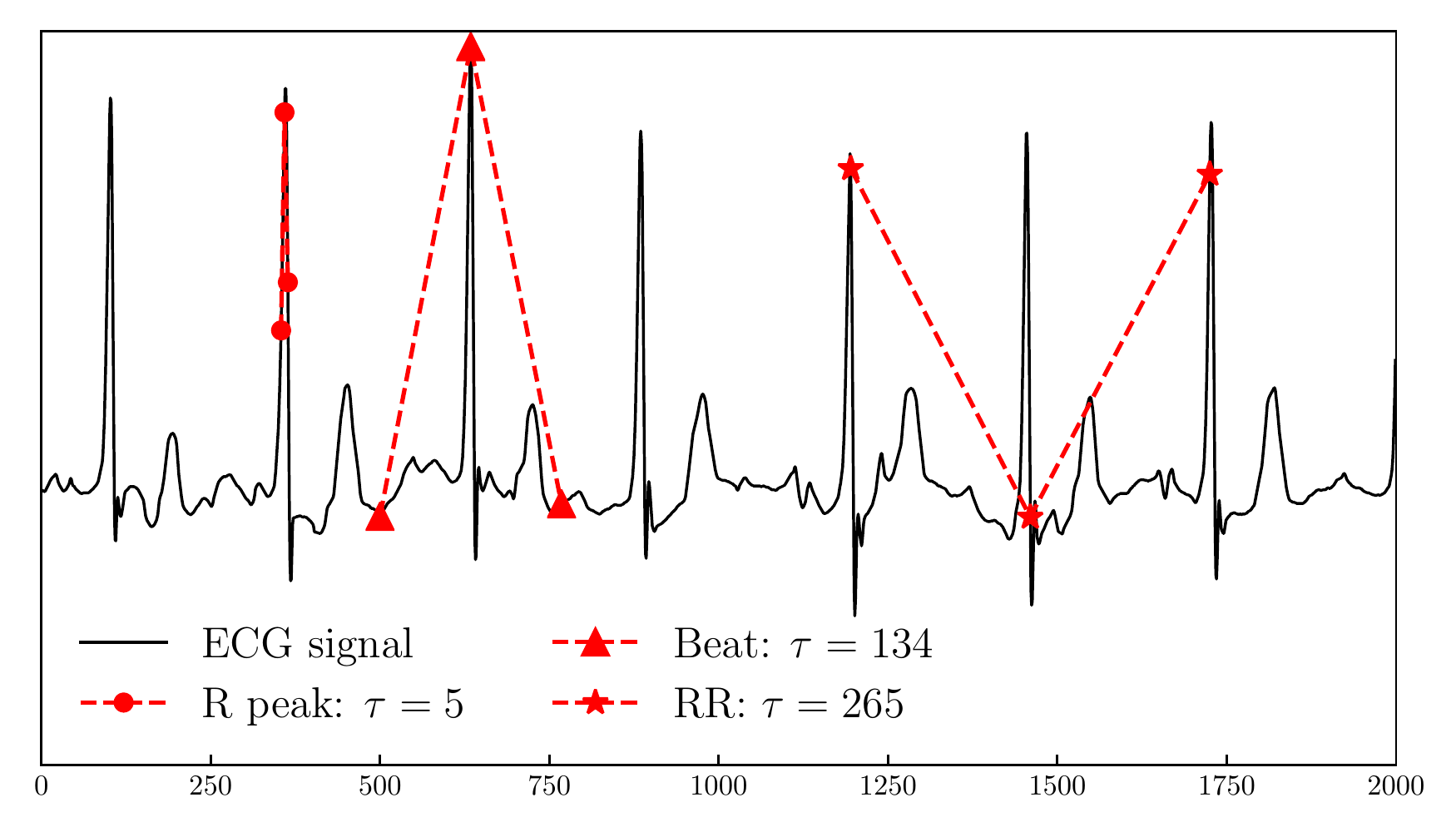}
\caption{The ECG signal and triadic time series motifs with different delays, $\tau$. $\tau=5$ corresponds to the R peak, $\tau=134$ beat and $\tau=265$ RR.}
\label{fig:triad}
\end{figure}

The time series motifs are the sub-sequences that are widely used in pattern recognitions in time series \cite{motif_2003,motif_2005,motif_2006}. 
Triadic Motif Field (TMF) is constructed based on triadic time series motifs.
Given a time series $\mathbold{X}=\{x(n), n=1,2,3,\cdots,N\}$, the sequence of triadic time series motifs is defined as 
\begin{equation}
  \mathbold{M}=\{\mathbold{M}(n|\tau), n=1,2,3,\cdots, N-2\tau\}
\end{equation}
where $\mathbold{M}(n|\tau)$ is a triadic time series motif $\mathbold{M}(n|\tau) = [x(n), x(n+\tau), x(n+2\tau)]^\top$ with delay $\tau$.
In the definition of a triadic time motif $\mathbold{M}(n|\tau)$, the delay $\tau$ is the separation step of two sequential points. The range of $\tau=1,2,3,\cdots,\tau_{max}$ is $\lfloor (N-1)/2 \rfloor$. Usually, $\tau$ is 1. However, the different delays are also important in detecting the long-range patterns in time series. Taking an ECG signal as an example, the three triadic time series motifs with different $\tau=5, 134, 265$ can approximately match the R peak, beat, and RR interval as shown in Fig.~\ref{fig:triad}. 

According to the triadic time series motifs with different delays, the TMF array $\mathbold{V}$, a three-channel image $\mathbold{V} \in \mathbb{R}^{\tau_{max} \times (N-2) \times 3}$, is defined as a stack of triadic time series motifs with all the possible delays, 
\begin{equation}
  \mathbold{V}_{\tau n k} = \mathbold{I}_k(n|\tau), k=1,2,3
\end{equation}
where
\begin{equation}
  \mathbold{I}(n|\tau) = 
  \begin{cases}
    \mathbold{M}(n|\tau),  &\; if\; 1 \leq n \leq N-2\tau \\
    [0, 0, 0]^\top , &\; if\; N-2\tau <n \leq N-2
  \end{cases}
\end{equation}

Finally, the TMF image, $\mathbold{TMF} \in \mathbb{R}^{\tau_{max} \times (N-2) \times 3}$, is defined as, 
\begin{equation}
  \mathbold{TMF}_{\tau n k} = \mathbold{V}_{\tau n k} + \mathbold{K}[\tau,n] \cdot \mathbold{V}_{\tau' n' k}
\end{equation}
where $\tau'=\tau_{max}-\tau+1$, $n'= N-n-1$ and $\mathbold{K} \in \mathbb{R}^{\tau_{max} \times (N-2)}$ is defined as a masker to prevent overlapping of the two arrays,
\begin{equation}
  \mathbold{K}[\tau,n] = 
  \begin{cases}
    0,  &\; if\; 1 \leq n \leq N-2\tau \\
    1 , &\; if\; N-2\tau < n \leq N-2
  \end{cases}
\end{equation}

\begin{figure}[H]
\centering
\includegraphics[width=0.5\textwidth]{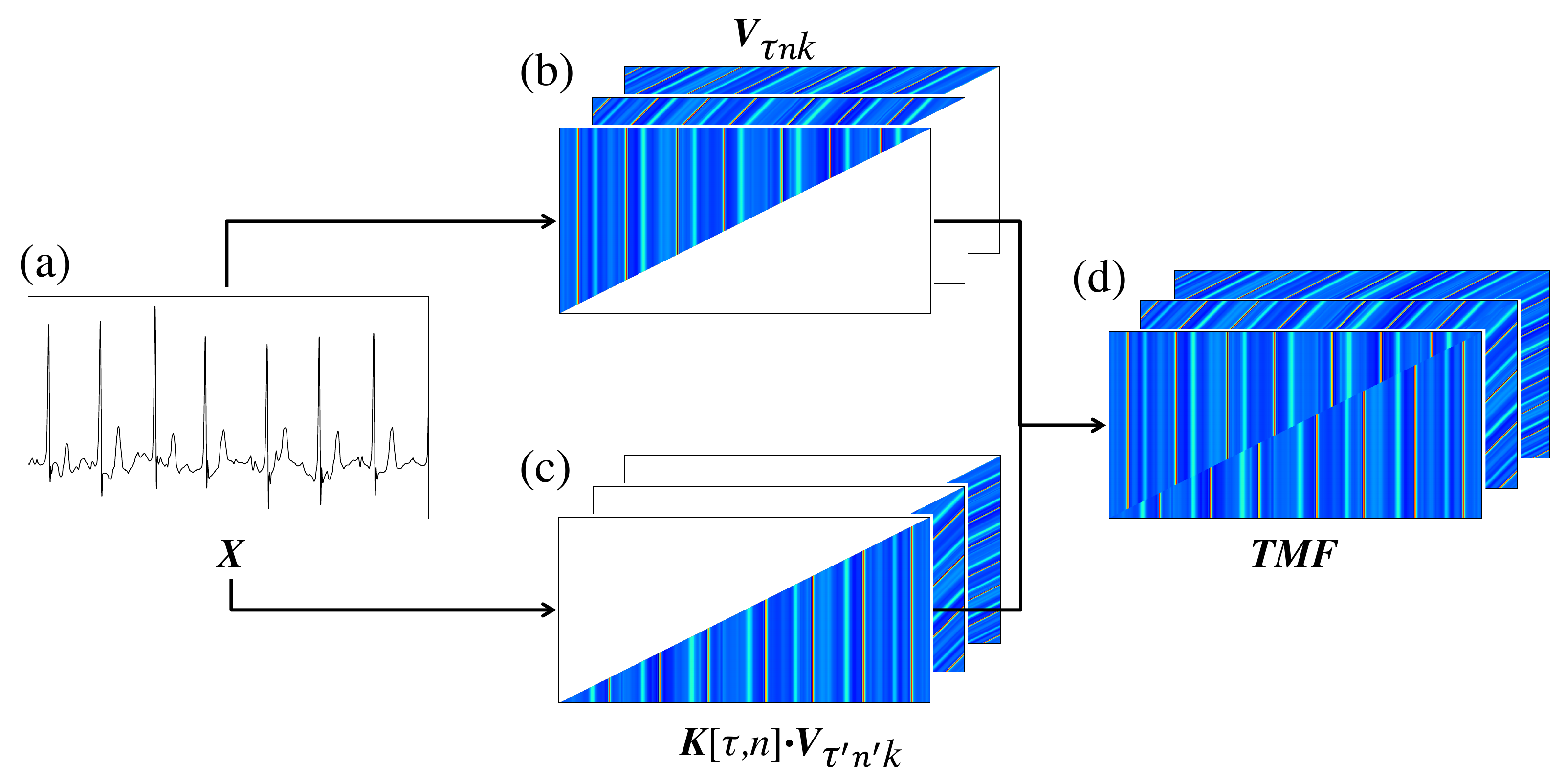}
\caption{The TMF images of an ECG signals. The blank spaces in (b) and (c) are 0 due to the structures of the TMF arrays $\mathbold{V}$. }
\label{fig:TMF}
\end{figure}

Fig.~\ref{fig:TMF} shows the procedure of encoding an ECG signal into a TMF image. In a word, the horizontal and vertical axises of TMF image are associated with the temporal and multi-scale informations of the triadic time series in the signal. 

\subsection{Deep Feature Transfer Learning and TMF Classification}\label{sec:classify}

With the transferred knowledge of instances, feature representations, and parameters, the deep transfer learning\cite{survey_tranfer_2010,transfer_2014} has been applied successfully in many real-world applications such as classification, regression, and clustering problems. The transfer learning pre-trained models are used to extract the features based on the TMF images of ECG signals. 

\begin{figure}[H]
  \centering
  \includegraphics[width=0.5\textwidth]{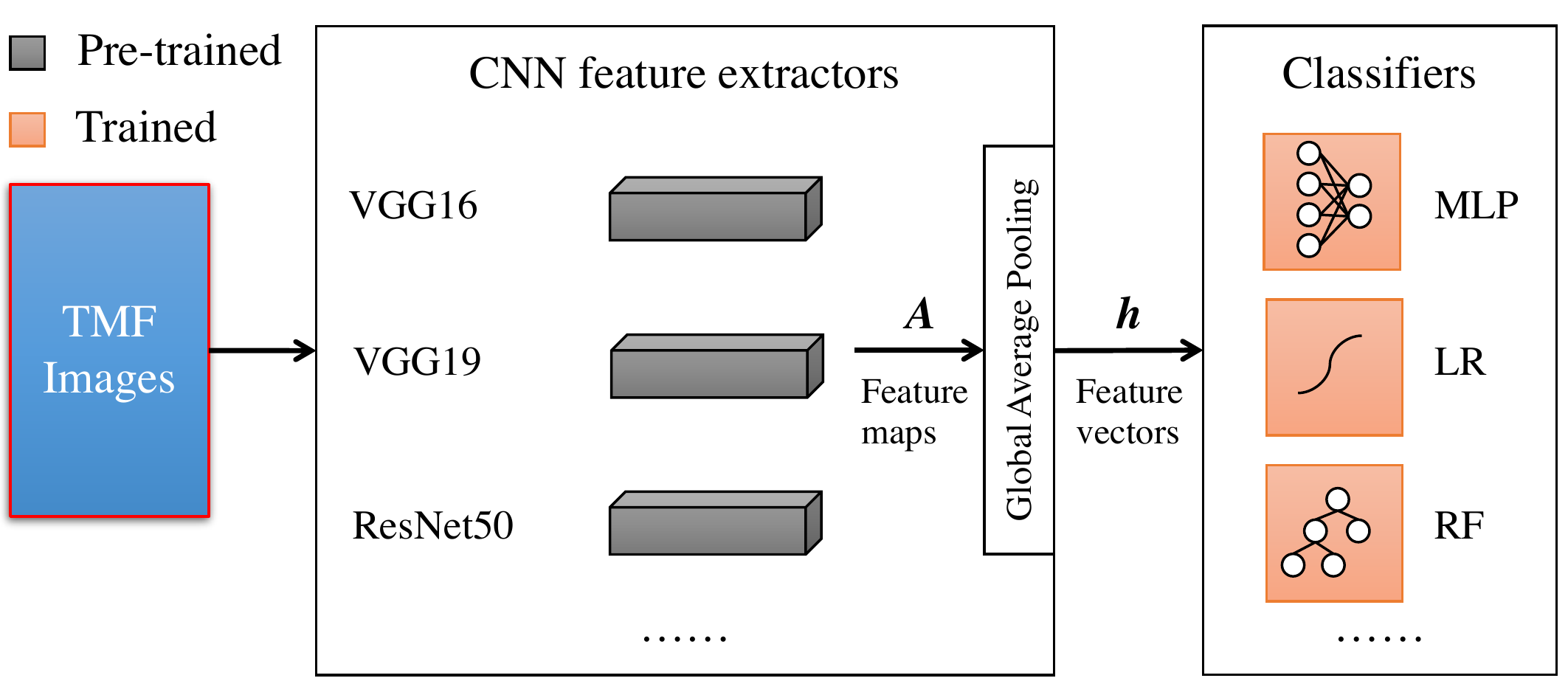}
  \caption{Architecture of the TMF classification model.}
  \label{fig:network}
\end{figure}

The TMF classification model has two parts, feature extractors and classifiers as shown in Fig.~\ref{fig:network}. The transfer learning feature extractors can be the pre-trained network models $\mathcal{F}$ including VGG16 \cite{vgg_2015}, VGG19 \cite{vgg_2015}, ResNet50 \cite{ResNet_2016}, and {\it etc.} The multi-channel feature map, $\mathbold{A} \in \mathbb{R}^{W\times H \times S}$, is extracted as,
\begin{equation}
  \mathbold{A} = \mathcal{F}(\mathbold{TMF})
\end{equation}
where $W, H, S$ are the width, height and channel number of the feature map. Followed by the Global Average Pooling (GAP) layer, the feature map $\mathbold{A}$ is aggregated as the feature vector,
\begin{equation}
  \mathbold{h} = \frac{1}{W\cdot H}[\sum_i^W\sum_j^H\mathbold{A}_{ij1}, \sum_i^W\sum_j^H\mathbold{A}_{ij2}, \cdots, \sum_i^W\sum_j^H\mathbold{A}_{ijS}]^\top
\end{equation}
With feature vector as the input, the classifiers can be multi-layer perceptron (MLP), logistic regression (LR), random forest (RF), and {\it etc}.
The MLP classifier includes one hidden layer with 128 neurons in (\ref{hidden}) and the predicted probability $\mathbold{y}$ of AF or non-AF is given with the last layer in (\ref{prob}),
\begin{equation}\label{hidden}
  \mathbold{h}_o = ReLU(\mathbold{W} \mathbold{h} + \mathbold{b})
\end{equation}
\begin{equation}\label{prob}
  \mathbold{y} = Softmax(\mathbold{W}_o \mathbold{h}_o + \mathbold{b}_o)
\end{equation}
where $\mathbold{y}$ is a $C$ dimension vector and $C$ is the number of classes. In this problem, $\mathbold{y}$ is $[y_1, y_2]$ where $y_1$ and $y_2$ indicate the AF and non-AF probabilities.

\subsection{Frame-wise Preprocessed Datasets}

The ECG recordings in PhysioNet Challenge 2017 database \cite{standard_database_2000,clifford_AF_2017} are used for the training and test of the classification models. The dataset contains ECG recordings of 8528 recordings sampled among which there are 738 AF patients and 7790 non-AF controls including normal, noisy, and other types. 

We divide the dataset according to the recordings into the training dataset (75\%), the validation dataset (10\%), and the test dataset (15\%) as shown in Table~\ref{tab:dataset-patient-AF}. Each ECG recording is converted into the sliding frames with a length of 3000. 50 and 500 strides are used for the sliding windows of the AF and non-AF ECG recordings. For the TMF classification model, the TMF images of the frames are used for the training and test. The preprocessed datasets consisting of the sliding frames are summarized in Table~\ref{tab:dataset-AF}.

\begin{table}[H]
  \centering
  \caption{Number of recordings in the preprocessed datasets for the AF patients and non-AF controls.}
  \begin{tabular}{cccc}
  \hline
  Type  &  Training & Validation & Test\\
  \hline
  AF patients      &   564    &  70   & 124  \\
  non-AF controls  &   5832    &  782   & 1156  \\
  \hline
  \end{tabular}
  \label{tab:dataset-patient-AF}
\end{table}

\begin{table}[H]
  \centering
  \caption{Number of frames in the preprocessed datasets for the AF and non-AF classes.}
  \begin{tabular}{cccc}
  \hline
  Type  &  Training & Validation & Test\\
  \hline
  AF frames      &   75979    &  8402   & 17317  \\
  non-AF frames  &   79423    &  10462   & 15865  \\
  \hline
  \end{tabular}
  \label{tab:dataset-AF}
\end{table}

The transfer learning models, VGG16, VGG19, and ResNet50 pre-trained on the ImageNet dataset excluding their top fully-connect layers in TensorFlow\cite{tensorflow} are used. The three classifiers in Fig.~\ref{fig:network} are trained, validated, and tested on the same datasets in Table~\ref{tab:dataset-AF}. 
With categorical cross entropy as loss function, the MLP classifier is trained with Adam \cite{Adam_2015} at the learning rate 0.001, $\beta_1$ 0.9 and $\beta_2$ 0.999, and it is validated with early stopping criteria on validation set.
Using Scikit-learn \cite{scikit-learn}, the hyper-parameters in LR and RF classifiers are tuned with the random-search strategy on the validation dataset. For the LR classifier, the selection of the penalty, solver, and {\it etc.} are optimized. For the RF classifier, the number of trees, number of features, and {\it etc.} are optimally selected. 
The system is equipped with HPC server, 32 dual CPU (2650 v3) nodes/640 cores. Our code will be publicly available later.

\section{Classification Performance and Discussion} \label{sec:benchmark}

The preprocessed datasets consist of the moving frames for the ECG signals of patients since the ECG recordings have varied lengths. 
The classification performance is measured by the area under the Receiver Operating Characteristic (ROC-AUC), Area under the Precision Recall Curve (PR-AUC), and the F1 score according to the sliding frames of patients in the test dataset. In the subsection~\ref{patient-wise} and subsection~\ref{clustering}, we will discuss the patient-wise classification performance and clusterings.

The TMF (VGG16-MLP) classification model is benchmarked with the three existing classification methods including the random forest with expert features (ExpertRF) \cite{mina_2019}, the convolutional recurrent network with spectrogram images (CRNN) \cite{CRNN_ECG_2017} and the multi-level attention network (MINA) \cite{mina_2019}. The VGG16 pre-trained model is used for feature extraction and the MLP classifier is trained with the early stopping criteria on the validation set. We then test it 5 times using different random seeds and report its mean values with standard deviations. 
Table~\ref{tab:result-AF} shows the TMF (VGG16-MLP) classification model outperforms all benchmarking models and has the F1 score 5.01\% higher than the current best model, MINA.

\begin{table}[H]
  \caption{Performance comparison on AF classification (\%)}
  \centering
  \begin{tabular}{cccc}
  \hline
  Method     &   ROC-AUC   &   PR-AUC  & F1  \\ 
  \hline
  ExpertRF   &   $93.94\pm0.00$    &  $88.16\pm0.00$   & $81.80\pm0.00$  \\
  CRNN       &   $90.40\pm1.15$    &  $89.43\pm1.11$   & $82.62\pm2.15$  \\
  MINA       &   $94.88\pm0.81$    &  $94.36\pm0.82$   & $83.42\pm2.29$  \\
  \hline
  TMF (VGG16-MLP) &   $\mathbf{95.50}\pm0.50$   &  $\mathbf{95.84}\pm0.42$   & $\mathbf{88.43}\pm0.39$\\
  \hline
  \end{tabular}
  \label{tab:result-AF}
\end{table}

Furthermore, Table~\ref{tab:result-net-classifier} shows the frame-wise performance for the TMF classification model with different feature extractors and classifiers. 
The pre-trained VGG16 \cite{vgg_2015}, VGG19 \cite{vgg_2015} and ResNet50 \cite{ResNet_2016} on the ImageNet dataset followed by Global Average Pooling are used as feature extractors. MLP, LR and RF are used as classifiers. 
LR and RF are selected with the random-search strategy on the validation dataset to tune the hyper-parameters. It shows that the VGG16-MLP combination has the best performance.

\begin{table}[H]
  \caption{Comparison of performances of three feature extractors, VGG16, VGG19 and ResNet50, and three classifiers, MLP, LR and RF.}
  \centering
  \begin{tabular}{cccc}
  \hline
  Model     &   ROC-AUC   &   PR-AUC  &  F1  \\ 
  \hline
  VGG16-MLP &  $\mathbf{95.50}\pm0.50$   &  $\mathbf{95.84}\pm0.42$   & $\mathbf{88.43}\pm0.39$\\
  VGG16-LR &   $95.49\pm0.00$   &  $95.64\pm0.00$   & $87.24\pm0.00$\\
  VGG16-RF &   $93.72\pm0.00$   &  $94.33\pm0.00$   & $83.07\pm0.00$\\
  \hline
  VGG19-MLP&   $95.21\pm0.31$   &  $95.38\pm0.31$   & $87.57\pm0.55$\\
  VGG19-LR &   $94.67\pm0.00$   &  $94.68\pm0.00$   & $87.74\pm0.00$\\
  VGG19-RF &   $92.50\pm0.00$   &  $93.07\pm0.00$   & $80.56\pm0.00$\\
  \hline
  ResNet50-MLP &  $93.85\pm0.31$   &  $93.92\pm0.46$   & $86.37\pm0.71$\\
  ResNet50-LR &   $94.40\pm0.00$   &  $94.33\pm0.00$   & $86.74\pm0.00$\\
  ResNet50-RF &   $89.35\pm0.00$   &  $90.51\pm0.00$   & $78.04\pm0.00$\\
  \hline
  \end{tabular}
  \label{tab:result-net-classifier}
\end{table}

\subsection{Frame Length of Test Dataset}

The frame-wise accuracy of the trained TMF (VGG16-MLP) classification model is also dependent on the frame length in the test. In the experiment, the VGG16-MLP model trained with 3000 frame length is evaluated on the test dataset where the frame length ranges from 100 to 3000 with step 100. 

\begin{figure}[ht]
  \centering
  \includegraphics[width=0.50\textwidth]{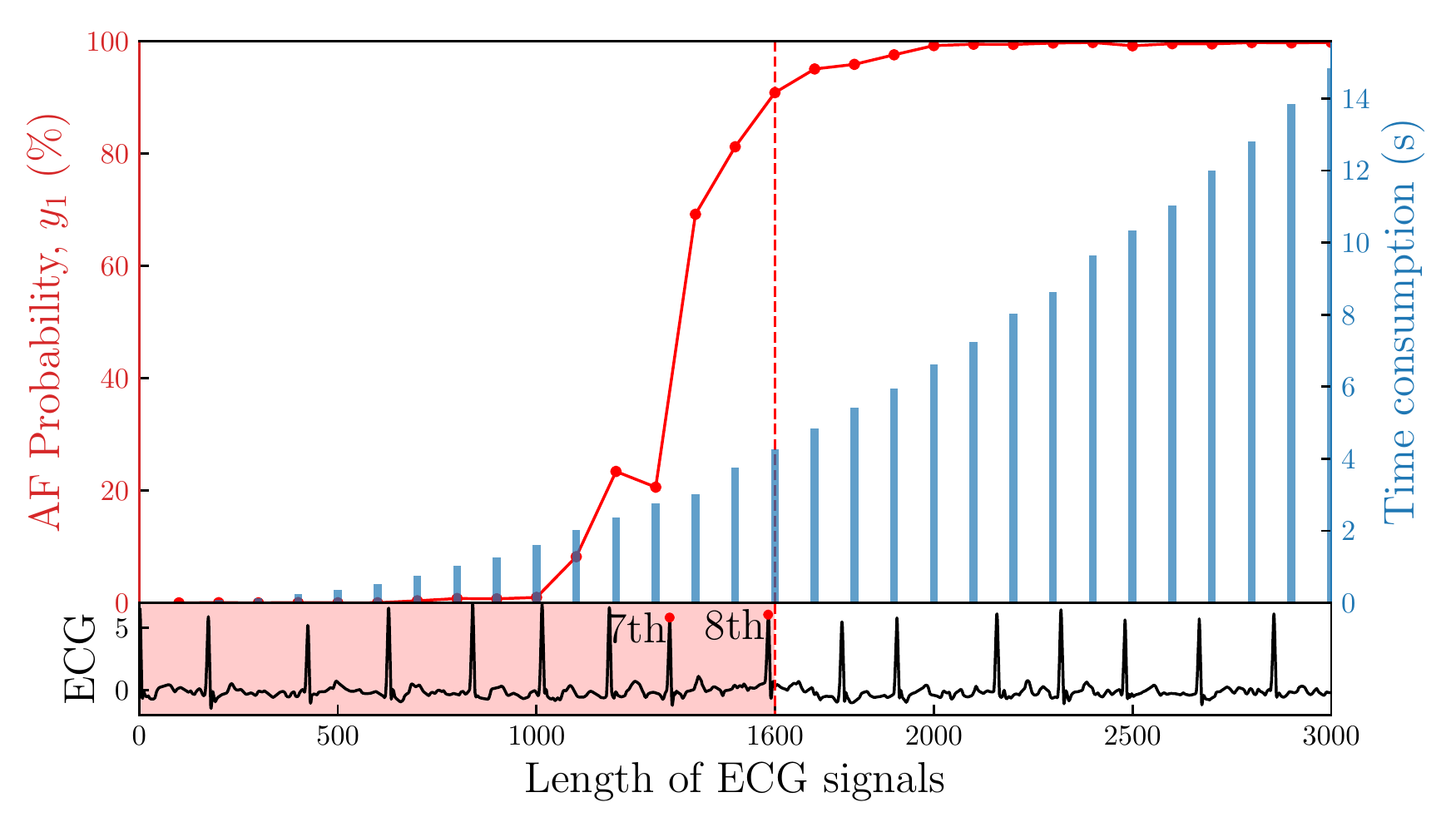}
  \caption{Length effect on the predicted AF probability for the ECG signal. Lower panel show an frame in the ECG signal. Upper panel shows the predicted AF probability $y_1$ and time consumption for the frame of different lengths.}
  \label{fig:flexibility}
\end{figure}

\begin{figure}[ht]
  \centering
  \includegraphics[width=0.50\textwidth]{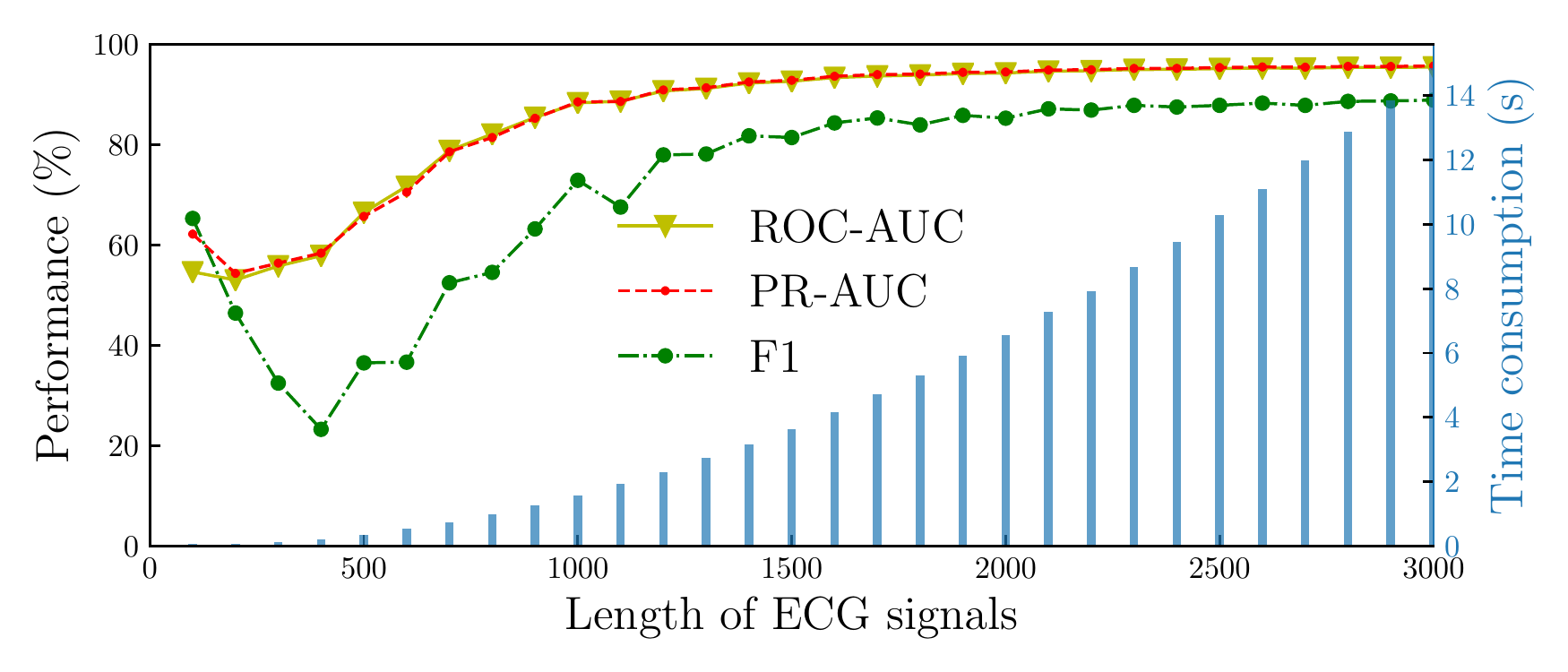}
  \caption{Frame length effect in the application of the classification model in the test dataset and the mean time consumed.}
  \label{fig:length_effect}
\end{figure}

With an AF ECG signal, as shown in Fig.~\ref{fig:flexibility}, we can see that the predicted AF probability $y_1$ increases rapidly when the length goes beyond the $7^{th}$ and $8^{th}$ R peaks corresponding to where the abnormal rhythm is located. With the frame 1600 step long,  the TMF classification model can predict the AF probability $y_1$ as well as the frame 3000 steps long. Therefore, the length of frames used in the test can be much shorter than the one in the training. As for statistical analysis in Fig.~\ref{fig:length_effect}, it is important to notice that the frame length longer than 1500 steps in the test dataset already can give a stable performance. Therefore, in the application of the TMF classification model, the frame 1500 step (5 seconds) long is sufficient enough. 

\subsection{Patient-wise Classification Performance}\label{patient-wise}
\begin{figure}[ht]
  \centering
  \includegraphics[width=0.50\textwidth]{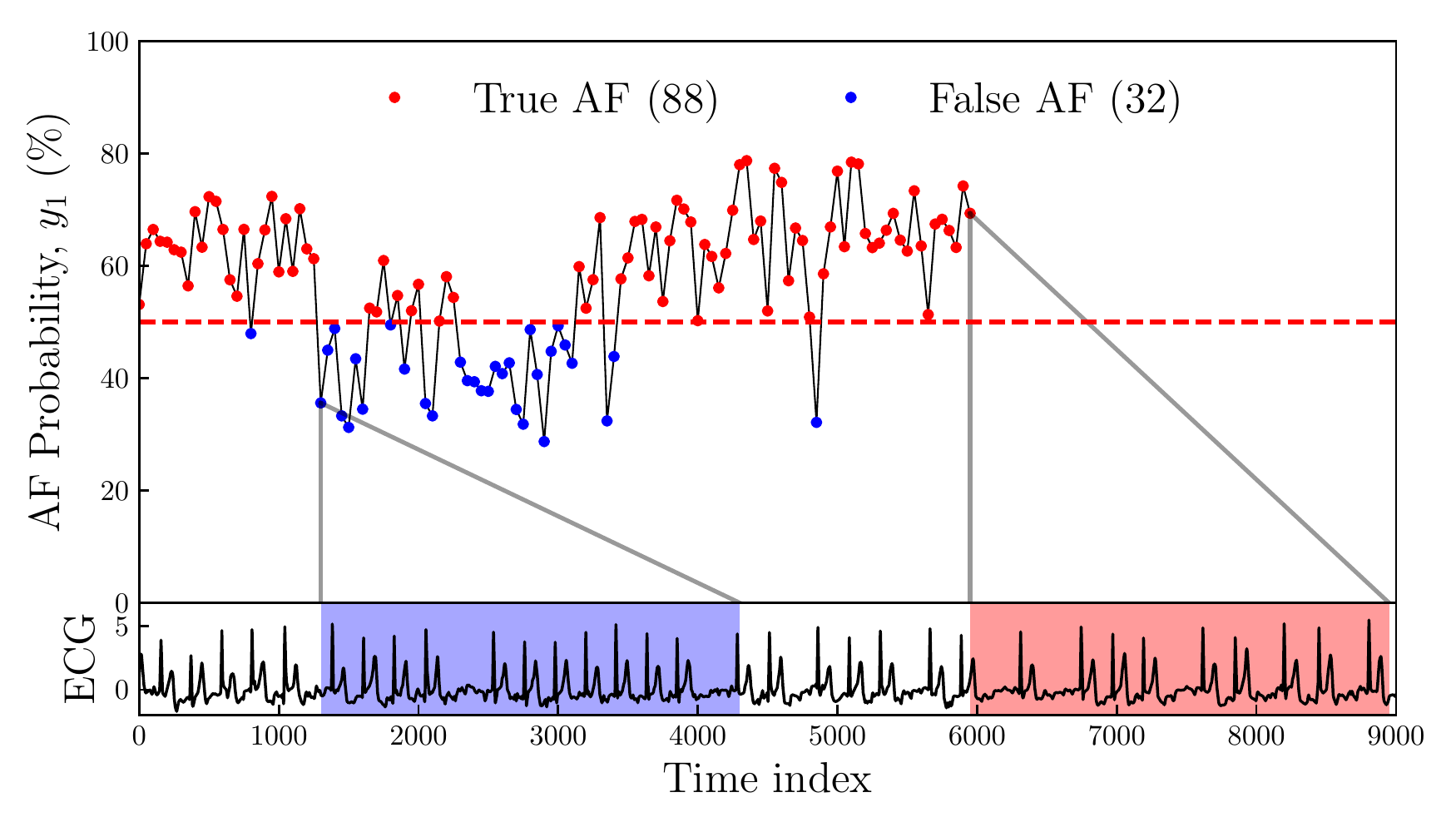}
  \caption{The classification probability ${y}_1$ for an AF patient. Lower panel is the full ECG signal of the patient. Upper panel indicates the predicted AF probability ${y}_1$ for the frames of the ECG signal. There are 88 frames predicted as AF class and 32 frames predicted as non-AF class in this patient. }
  \label{fig:patient-segment}
\end{figure}

Given the preprocessed dataset, the performance of the TMF classification model is evaluated with the frame-wise preprocessed datasets. 
However, the patient-wise classification is important for the clinical application. According to the frames belonging to one AF patient, the patient-wise classification accuracy is defined as,
\begin{equation}
  accuracy = \frac{m}{M} \times 100 \%
\end{equation}
where $m$ is the number of frames with which the TMF classification model predicts that it is the AF patient if the AF probability $y_1$ is higher than 50\%. $M$ is the total number of frames belonging to this patient.  For example, in Fig.~\ref{fig:patient-segment}, its patient-wise accuracy is $88/120=73.3\%$.

\begin{figure}[H]
  \centering
  \includegraphics[width=0.50\textwidth]{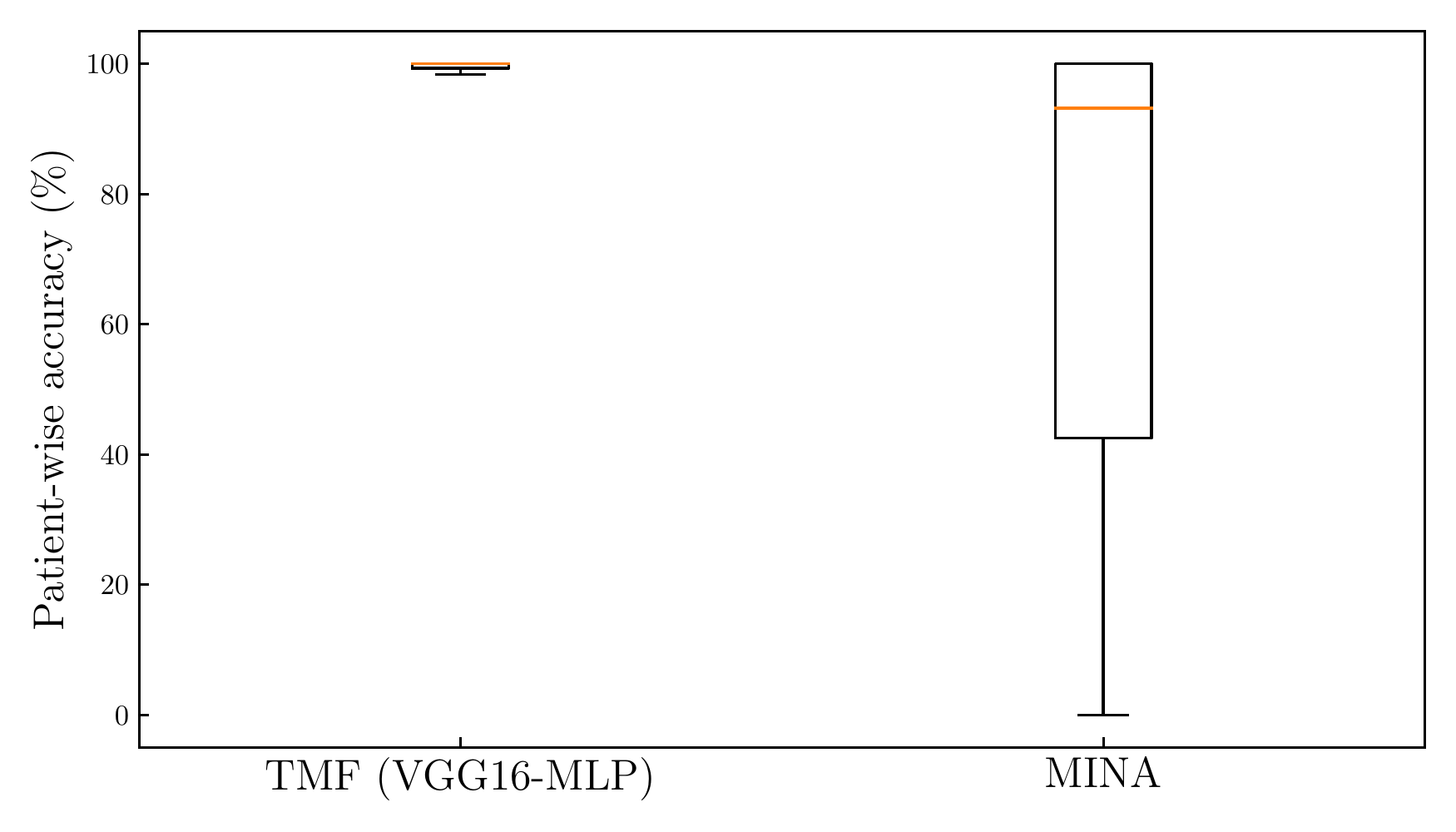}
  \caption{Box plot shows the patient-wise accuracy distribution according to the TMF classification model and MINA.}
  \label{fig:patient-box-plot}
\end{figure}

Fig.~\ref{fig:patient-box-plot} shows that the TMF (VGG16-MLP) classification model has better patient-wise accuracies and narrow distribution compared to MINA. According to the definition of the patient-wise accuracy, we define two extreme groups of the patient-wise classification results, totally incorrect (TI) and totally correct (TC). The TI and TC groups refer to the patients with the patient-wise accuracy equal to 0\% and 100\% respectively. In Table~\ref{tab:result-patient-classifier}, it shows that the TMF classification model can correctly identify 90 patients as AF among all 124 AF patients in the test dataset, which is 44 more than MINA.

\begin{table}[H]
  \caption{The number of patients in the totally incorrect (TI) and totally correct (TC) groups.}
  \centering
  \begin{tabular}{ccc}
  \hline
  Model     &   TI & TC\\
  \hline
  TMF (VGG16-MLP) &  $\mathbf{1}$ &   $\mathbf{90}$\\
  MINA        &  7 &   46\\
  \hline
  \end{tabular}
  \label{tab:result-patient-classifier}
\end{table}

\subsection{Patient-wise ECG Feature Clustering}\label{clustering}
\begin{figure*}[th]
  \centering
  \subfloat[]{\includegraphics[width=0.50\textwidth]{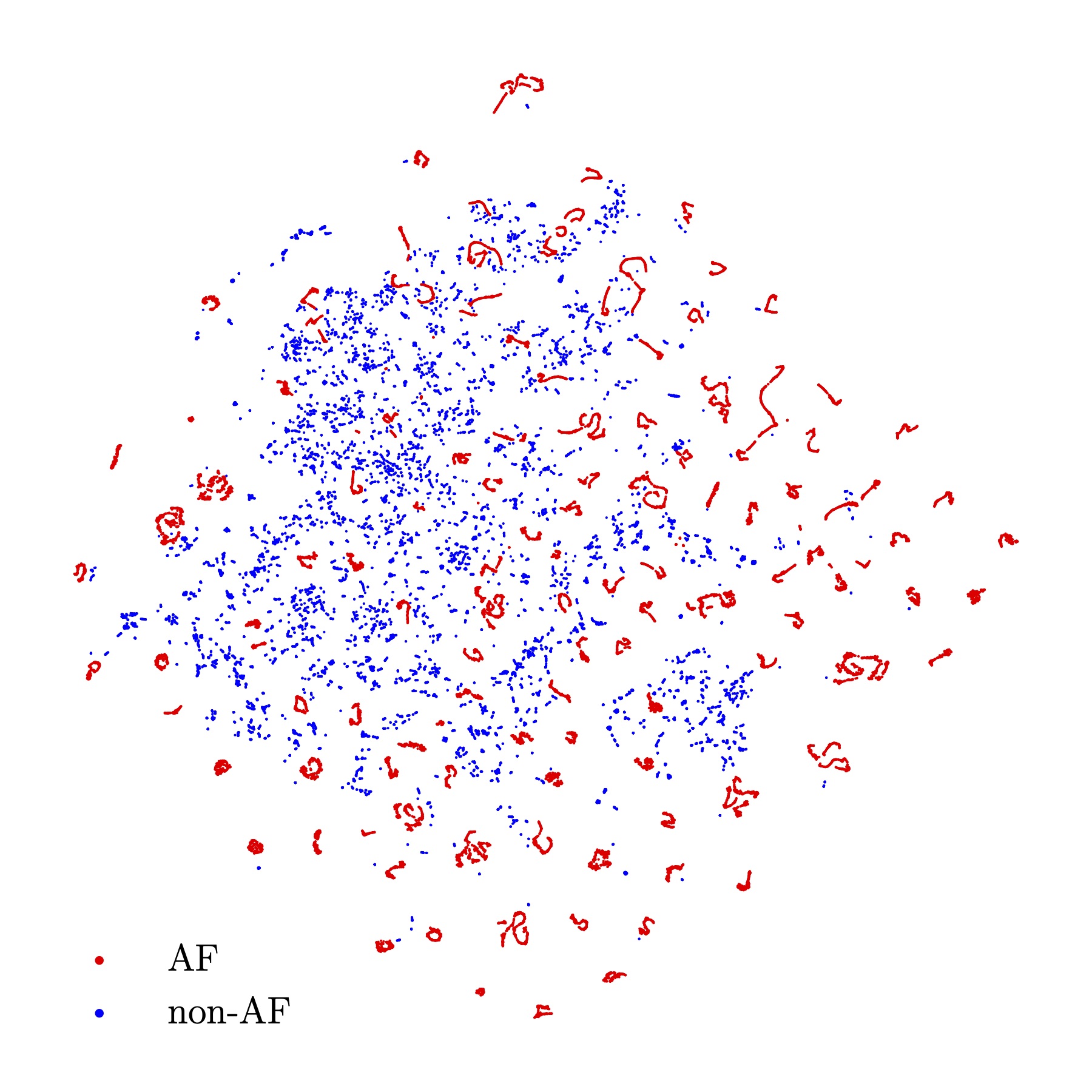}\label{fig:tsne-vgg}} 
  \subfloat[]{\includegraphics[width=0.50\textwidth]{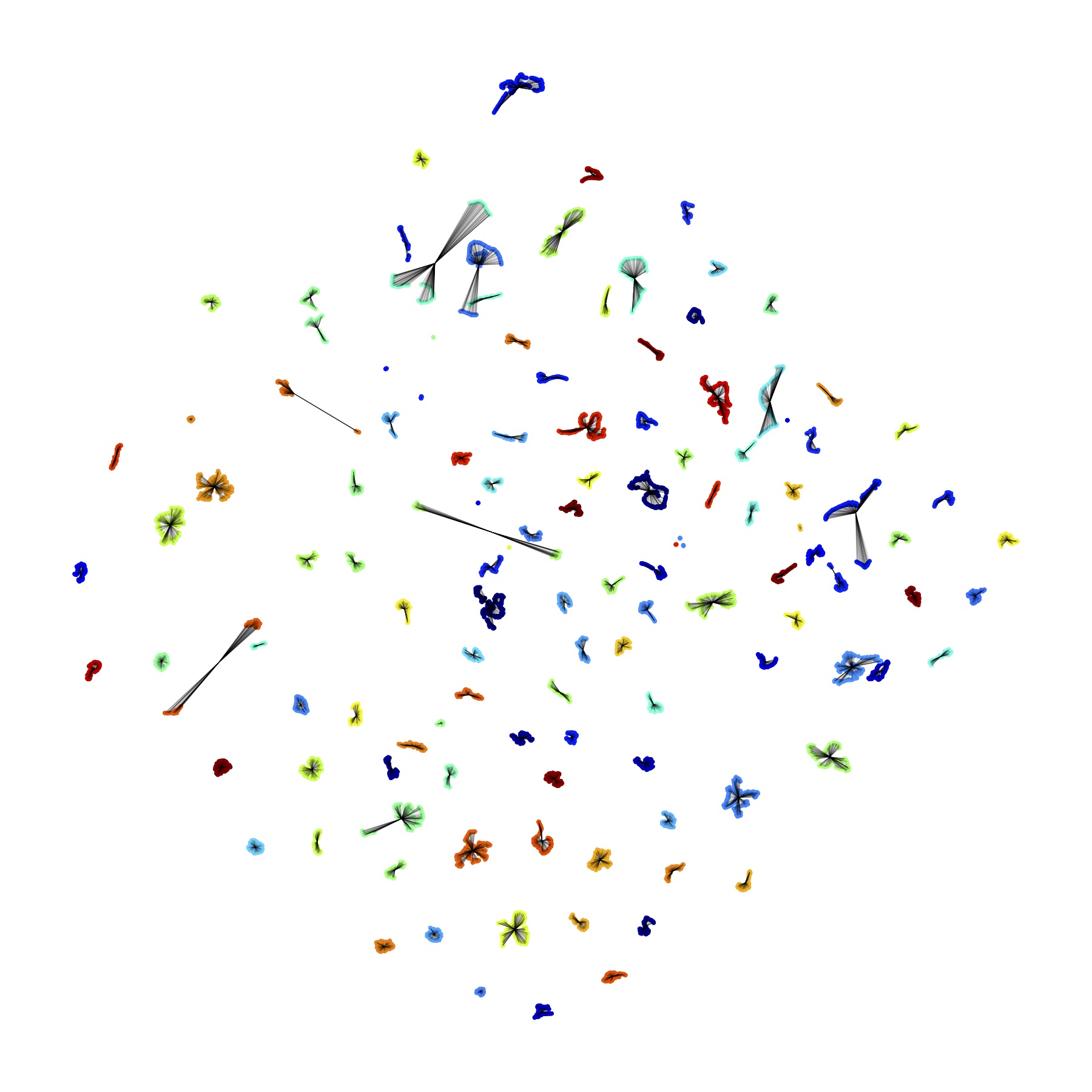}\label{fig:tsne-af-vgg}}
  \caption{Visualization of features extracted from TMF images on test dataset. (a) AF and non-AF clustering based on the frame-wise feature vectors. (b) Patient-wise clustering of the AF patients. All the frame-wise features belonging to the same patients are connected and painted  with the same colors.} \label{fig:tsne}
\end{figure*}

In order to visualize the transfered representation, t-distributed Stochastic Neighbor Embedding (t-SNE) \cite{tsne_2008} is used to map high-dimensional feature vectors $\mathbold{h}$ extracted 
Fig.~\ref{fig:tsne-vgg} shows the points in 2D space demonstrate clear separation between AF and non-AF classes of the test dataset. By repainting the AF points with the same color and connecting them according to the patients, Fig.~\ref{fig:tsne-af-vgg} shows clear patient-wise clusterings. Due to the knowledge transferred from ImageNet, the features naturally retain the basic similarity among frames in individual patients. 
Therefore transfer learning based on VGG16 performs well in the patient-wise feature extraction and classification as shown in Fig.~\ref{fig:patient-box-plot} and Table~\ref{tab:result-patient-classifier}.

\section{Image-based Temporal Pattern Recognition and Clinical Interpretation}
\label{sec:interpret}

The symmetrized Grad-CAM of the TMF images is used to interpret the TMF classification model for the ECG signals. The Gradient-weighted Class Activation Mapping (Grad-CAM) \cite{Grad_CAM_2017} is a technique to produce a visual explanation to varieties of CNN models. In the paper, we apply Grad-CAM to detect the ECG clinical temporal patterns in the TMF classification model. 

Given the TMF images of the ECG signal of the AF or non-AF class, the Grad-CAM, $\mathbold{L}^c\in \mathbb{R}^{W\times H}$ of the VGG16-MLP in Fig.~\ref{fig:network} can be generated by the gradient-based weighted combinations of the feature maps $\mathbold{A}$,
\begin{equation}
  \mathbold{L}^c = ReLU[\sum_s^S(\alpha^c_s \mathbold{A}_s)]
\label{equ:grad-cam-1}
\end{equation}
where $s$ is channel index of the feature map, $c$ is 1 or 2 that indicates AF or non-AF, and
\begin{equation}
  \alpha^c_s=\frac{1}{W\cdot H}\sum_i^W\sum_j^H\frac{\partial y_c}{\partial \mathbold{A}_{ijs}}.
\label{equ:grad-cam-2}
\end{equation}
Then, the Grad-CAM image will be up-sampled into $\mathbold{L}_{in}^c \in \mathbb{R}^{\tau_{max}\times (N-2)}$, which have the same width and height of the input TMF images. 

Due to the symmetry in the TMF images as constructed, we enforce the same symmetry by defining the new symmetrized Grad-CAM $\mathbold{G}^c_{\tau n}$ with $(\mathbold{L}_{in}^c[\tau, n]+\mathbold{L}_{in}^c[\tau',n'])/2$ where $(\tau,n)$ and $(\tau',n')$ are the two array indices according to the 180 rotation symmetry in $\mathbold{V}$ and the masked one,
\begin{equation}
  \mathbold{G}^c[\tau, n] = 
  \begin{cases}
  \mathbold{L}_{in}^c[\tau, n] , \; if\; \mathbold{K}[\tau, n] = \mathbold{K}[\tau', n'] =0 \\
  (\mathbold{L}_{in}^c[\tau, n] + \mathbold{L}_{in}^c[\tau', n'])/2,\; others \;
  \end{cases}
\end{equation}
where $\tau'=\tau_{max}-\tau+1$ and $n'=N-n-1$. 
$\mathbold{G}^c$ gives the significance of the triadic motif in the TMF images of patient class $c$ of AF or non-AF. This helps us to identify the clinically meaningful patterns in the ECG signals.

\begin{figure*}[th]
  \centering
  \subfloat[AF frame: $\mathbold{G}^{1}$ with ${y}_{1}=94.03\%$]{\includegraphics[width=0.50\textwidth]{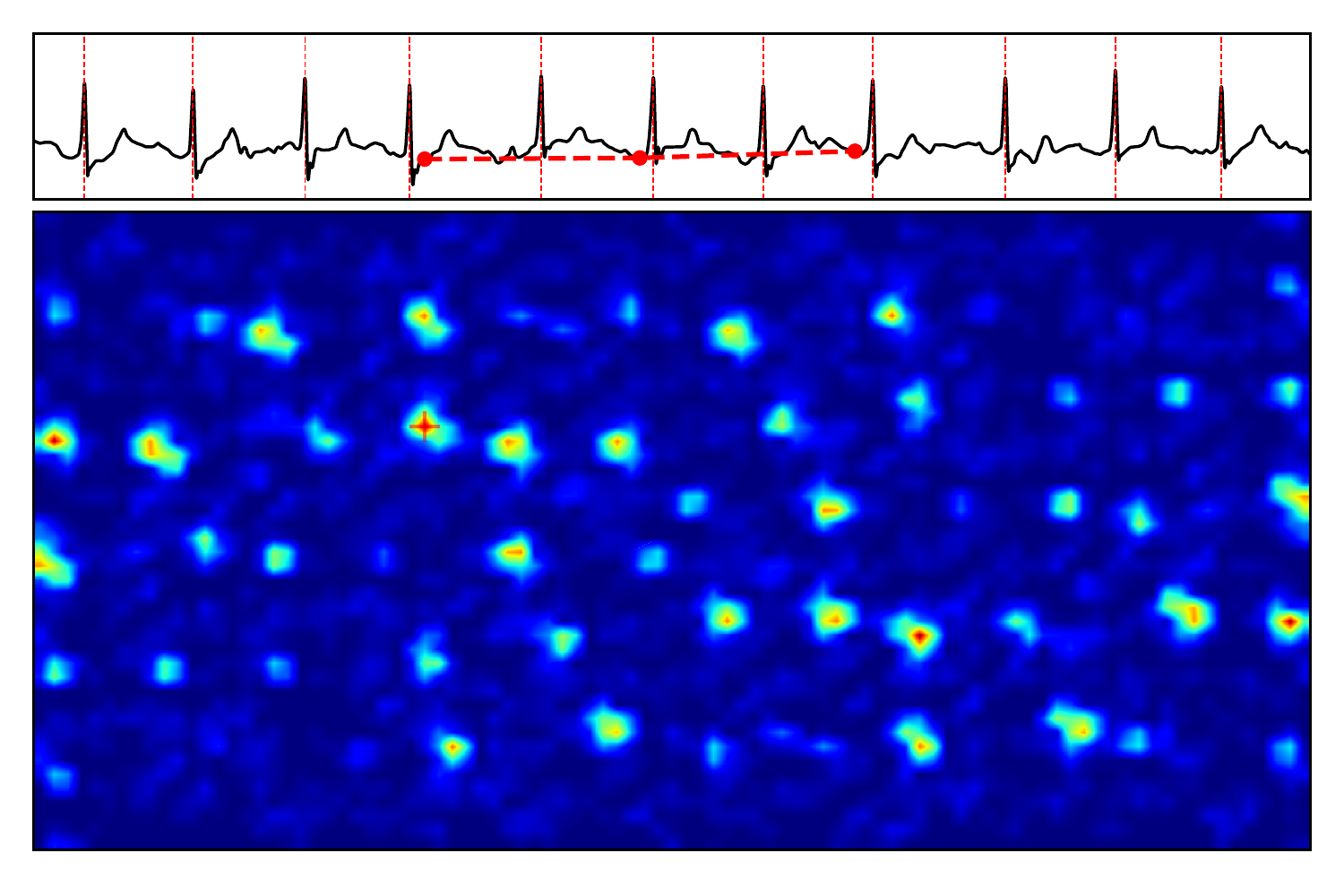}\label{fig:AF-AF}}
  \subfloat[Non-AF frame: $\mathbold{G}^{1}$ with ${y}_{1}=0.96\%$]{\includegraphics[width=0.50\textwidth]{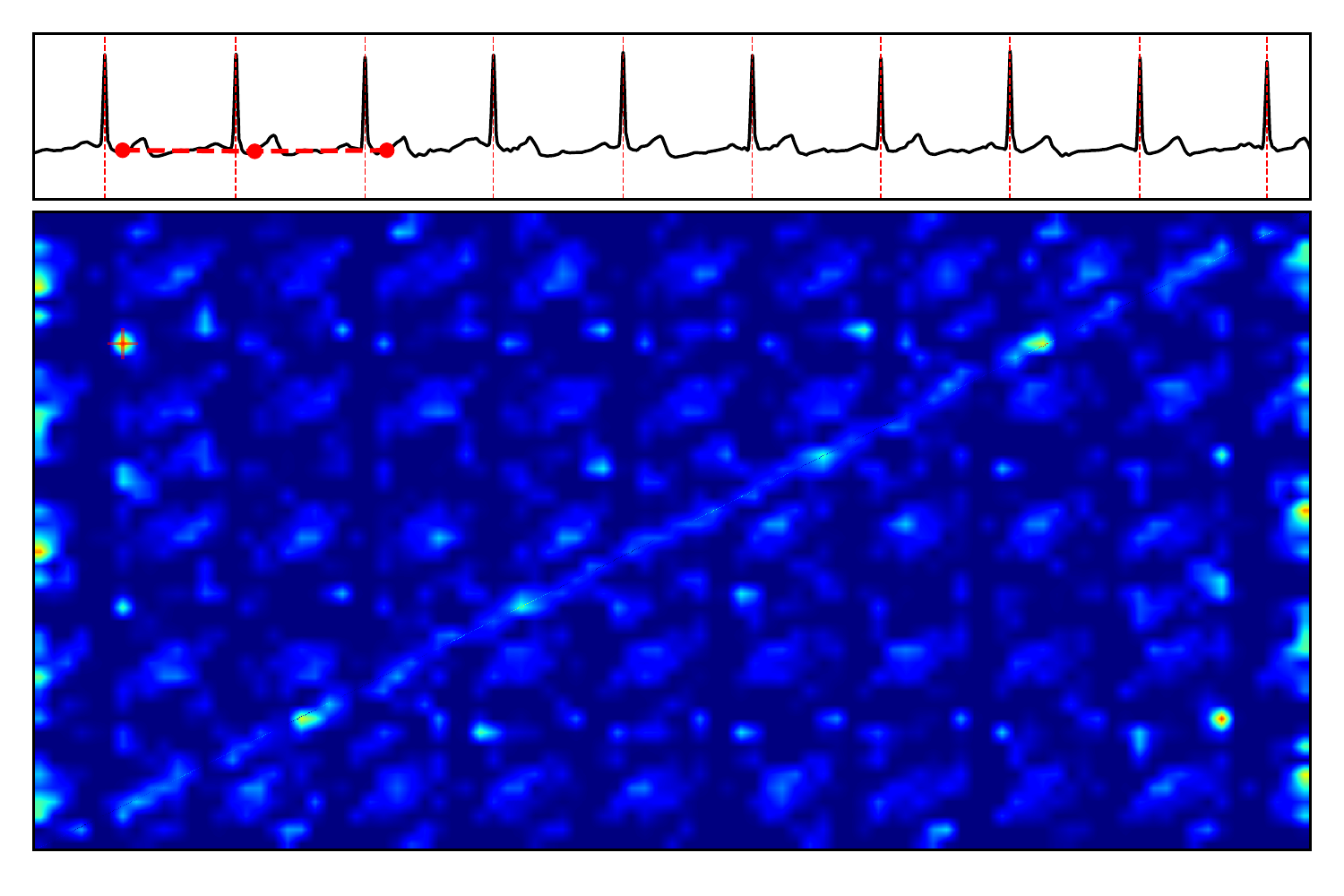}\label{fig:nAF-AF}}
  \\  
  \subfloat[AF frame: $\mathbold{G}^{2}$ with ${y}_{2}=5.97\%$]{\includegraphics[width=0.50\textwidth]{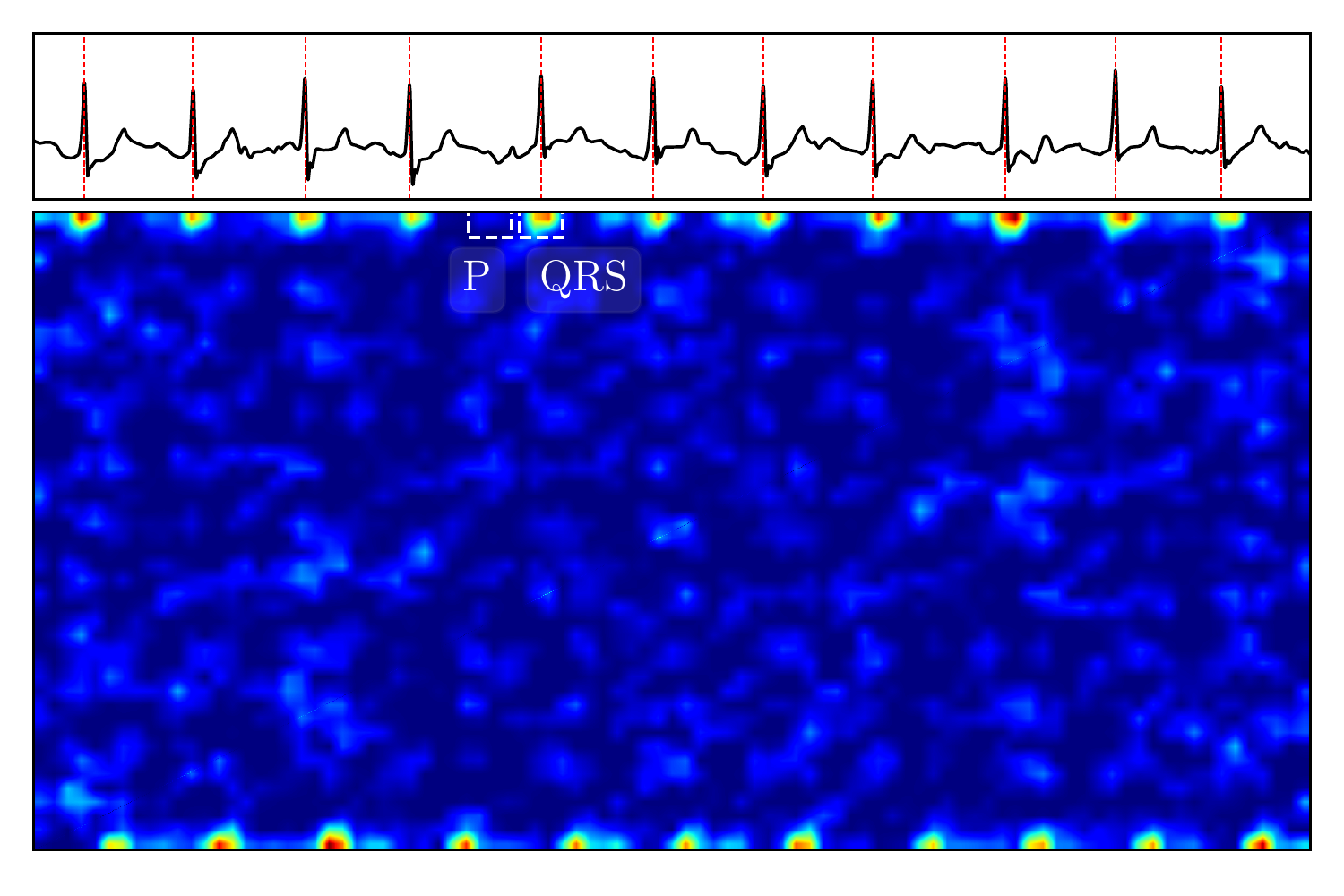}\label{fig:AF-nAF}}
  \subfloat[Non-AF frame: $\mathbold{G}^{2}$ with ${y}_{2}=99.04\%$]{\includegraphics[width=0.50\textwidth]{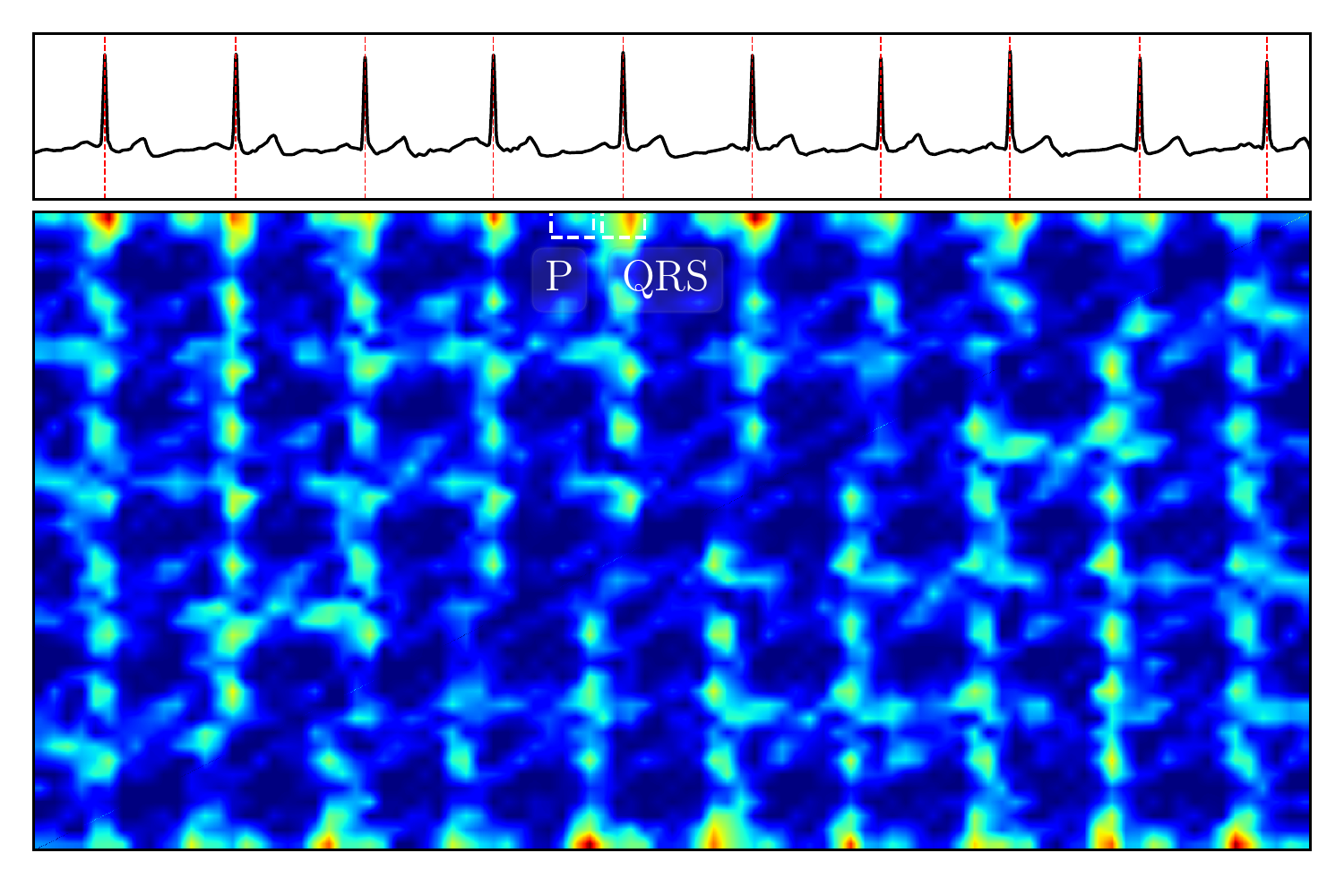}\label{fig:nAF-nAF}}
  \\ 
  \caption{The AF and non-AF signals and their symmetrized Grad-CAM images. (a) and (c) show the symmetrized Grad-CAM images of the AF ECG frame while (b) and (d) show the non-AF ECG frame. Vertical dash lines in ECG signals indicate the R peaks. The triadic time series motifs with big delay associated abnormal non-periodic interval in the ECG recording are labeled with the red crosses in symmetrized Grad-CAM images in (a) and (b).} \label{fig:localization:AF}
\end{figure*}

Cardiologists diagnose an ECG recording as the AF or non-AF case according to the following clinical patterns: 1) P wave; 2) RR interval. 
The TMF images consist of triadic time series with different delay and initial time indices that are associated with the clinical patterns. As shown in Fig.~\ref{fig:triad}, triadic motifs with small delays can match the beat-level patterns such as QRS complex and R peak, while the triadic motifs with big delays catch rhythm-level patterns such as RR interval. The initial time index of the triadic motif will locate the pattern temporally.

The symmetrized Grad-CAM demonstrates strong interpretability for the AF and non-AF recordings at the beat and rhythm levels. 
As shown in Figs.~\ref{fig:AF-nAF} and~\ref{fig:nAF-nAF}, hot spots with small delays are strongly associated with the QRS complex in the non-AF controls. In Fig.~\ref{fig:AF-nAF}, the weak regions on the left of the QRS complex indicate the abnormal or the absence of P waves in the AF recording while the strong and wide regions indicate the normal P waves in Fig.~\ref{fig:nAF-nAF}.
As shown in Fig.~\ref{fig:AF-AF}, the highlighted regions corresponding to the triadic motifs with big delays match the abnormal interval in the AF patient. 
In Figs.~\ref{fig:nAF-AF} and~\ref{fig:nAF-nAF}, the hot red spots demonstrate the spatial periodicity that naturally associated with the temporal periodicity of beats in the non-AF recording, while the non-periodicity in the AF recording shown in Figs.~\ref{fig:AF-AF} and~\ref{fig:AF-nAF} reflect the abnormal interval at the rhythm level. 
In summary, the symmetrized Grad-CAM can precisely match the clinically important patterns in the AF class at the beat and rhythm levels and distinguish the AF and non-AF classes.

\section{Concluding Remarks}

In this paper, the interpretable classification model based on TMF images is proposed to detect the Atrial Fibrillation in the ECG recordings. 
It outperforms the baseline models in the AF classification. The TMF image encoding scheme has the following advantages:
\begin{enumerate}
  \item The TMF images can visualize the order patterns and temporal structures of time series associated with the ECG analysis.
  \item The TMF (VGG16-MLP) classification model has state-of-the-art performance compared to the baseline models based on the moving frames. Besides, the TMF classification model has much better patient-wise accuracies and clusterings in terms of the moving frames belonging to one patient.  
  \item The TMF classification model is simple and effective to be implemented and doesn't cost too much time on training, model selection, and hyper-parameters optimization.
  \item The symmetrized Grad-CAM technique can be used to identify the clinical patterns in the TMF images of AF patients. It allows us to interpret the results from beat and rhythm levels. 
\end{enumerate}
The temporal data extensively exist in the medical, mechanical, electronic, and power systems. In the next step, we will further explore the applications of the TMF classification model in engineering systems such as fault detection, system health monitoring, distributed event detection in sensor networks, and {\it{etc}}. Currently, mobile edging computing devices such as smartphones all have good GPU and image processing capability. Given the advantages and efficiency of the TMF imaging encoding scheme, the TMF classification model has promising applications in edge computing and the Internet of Things (IoT).

\bibliographystyle{unsrt}
\bibliography{paper}

\end{document}